\documentclass[sigconf,nonacm]{acmart}
\AtBeginDocument{%
  }

\setcopyright{acmlicensed}
\copyrightyear{2018}
\acmYear{2018}
\acmDOI{XXXXXXX.XXXXXXX}
\acmConference[Conference acronym 'XX]{Make sure to enter the correct
  conference title from your rights confirmation email}{June 03--05,
  2018}{Woodstock, NY}
\acmISBN{978-1-4503-XXXX-X/2018/06}

\usepackage{makecell}

\begin{document}

\title{Unmasking Propaganda: A Comparative Analysis of Masked and Causal Language Models}

\author{Claudiu Creanga}
\email{claudiu.creanga@s.unibuc.ro}
\author{Ioachim Lihor}
\email{ioachim.lihor@gmail.com}
\author{Liviu P. Dinu}
\email{liviu.p.dinu@gmail.com}

\affiliation{%
  \institution{Faculty of Mathematics and Informatics, 
Human Language Technologies Research Center, Interdisciplinary School of Doctoral Studies, University of Bucharest}
    \country{Romania}
}

\renewcommand{\shortauthors}{Creanga et al.}

\begin{abstract}
Propaganda detection is an essential task in natural language processing (NLP), particularly in the context of manipulative political communications.  However, identifying specific propaganda techniques presents a significant challenge due to their often subtle nature and reliance on context, making them difficult to distinguish from legitimate persuasive language. Propaganda often involves highlighting certain facts while downplaying or ignoring others to create a desired perception. This biased communication aims to influence attitudes, beliefs, or behaviors towards a particular cause or position. This paper explores advances in detecting propaganda techniques through a comparative analysis of modern language models, using the SemEval-2020 Task 11 dataset. We evaluated both masked language models (based on XLM-RoBERTa or DeBERTa V3) and causal models (from OpenAI, Google, Mistral, Anthropic and Meta), employing two prompting strategies: base and chain-of-thought prompting. Our results demonstrate improvements over state-of-the-art models, with the best-performing MLM achieving an F1 score of 63.18 in technique classification and the best causal model achieving 63.62. We also observed that certain models excel in specific techniques, such as loaded language and name-calling, while struggling with others like bandwagon and black-and-white fallacy. These findings suggest that fine-tuning, ensemble modeling, and the use of larger datasets can further enhance propaganda detection capabilities.
\end{abstract}



\maketitle

\section{Introduction}

Propaganda represents a systematic and deliberate operation extending far beyond mere persuasion, aiming instead to actively manipulate public opinion and shape collective perception through the stringent control of information. This is achieved by strategically and selectively presenting narratives, meticulously highlighting facts, interpretations, or images that bolster a specific agenda, while simultaneously downplaying, omitting, distorting, or even fabricating information that contradicts it. This biased communication constructs a distorted reality in the minds of the audience, ultimately seeking to influence their attitudes, solidify specific beliefs, and steer their behaviour towards a predetermined cause, ideology, or political position.

Historically, propaganda has long served as an especially insidious and powerful instrument, particularly favored by authoritarian regimes seeking to maintain control (\cite{Stanley2015}). It proves invaluable for such powers, enabling them to manufacture consent, legitimize their grip on political authority (often by exaggerating achievements or inventing threats), and systematically stifle dissent by dominating the informational landscape and silencing or discrediting opposing viewpoints. By disseminating biased information, selectively framing events, leveraging potent emotional appeals over factual discourse, and sometimes spreading outright disinformation, propaganda crafts this distorted reality. This manufactured reality serves a dual purpose: it bolsters the authority and preferred narratives of the elite while simultaneously undermining the citizenry's ability to critically evaluate information and challenge the status quo.

To achieve these manipulative goals, propagandists employ a diverse and often sophisticated toolkit. Techniques range from the overt use of emotionally charged or loaded language, derogatory name-calling to demonize opponents, and the incessant repetition of slogans or messages until they are widely accepted, to more subtle tactics. These can include the oversimplification of complex issues into misleading dichotomies, potent emotional appeals designed to bypass rational analysis, concerted efforts to sow doubt about credible alternative sources, gross exaggeration or minimization of events, and the clever exploitation of logical fallacies to create a superficial veneer of reasoned argument. The corrosive effects of propaganda on society are multifaceted. They lead to heightened polarization and social division, frequently employing 'us vs. them' tactics to create rifts within populations. Furthermore, propaganda contributes to a decline in critical thinking as citizens are encouraged to passively accept official narratives rather than engage in independent analysis. A particularly dangerous consequence is the dehumanization of designated out-groups, which can be used to justify mistreatment or marginalization.

This landscape of manipulation exists within a broader context of challenges to information integrity. There is substantial evidence of a worldwide crisis in news trust, prompting significant debate on how to restore public confidence (\cite{FlewEtAl2020}). This crisis is unfolding as news consumption patterns shift dramatically, with more people, particularly the young, increasingly obtaining news online, and many others choosing to avoid news altogether. Consequently, within the complex and rapidly evolving digital information ecosystem—where information spreads virally and can be targeted with precision—the task of identifying, analyzing, and exposing propaganda has become fundamental for preserving informed public discourse and democratic health (\cite{DaSanMartinoEtAl2020b}). 

In this research, we used Masked Models and Causal Models to obtain good results in detecting propaganda techniques using the SemEval 2020 dataset. We make our code publicly available on GitHub. 

\section{Related Work}
Propaganda has not been as extensively studied in NLP as other topics because of the complexity of detection, lack of labeled data and subjectivity. The first dataset that we are aware of is TSHP-17 created by \cite{rashkin-etal-2017-truth} which annotated documents in 4 classes: trusted, satire, hoax and propaganda. Another dataset, Qprop \cite{BARRONCEDENO20191849}, was released in 2019 and contained a binary imbalanced dataset with ~10\% of documents belonging to propaganda. PTC \cite{DBLP} was released in the same year and contained specific text spans that were labelled according to 18 propaganda techniques. In 2020, at SemEval task 11: Detection of propaganda techniques in news articles \cite{da-san-martino-etal-2020-semeval} a dataset was presented with 2 sub-tasks: Span Identification (SI) and Technique Classification (TC). The Span Identification sub-task focuses on locating specific segments of text that employ propaganda techniques. Once these segments are identified, the Technique Classification sub-task categorizes the particular propaganda methods used within those spans. This approach of 2 sub-tasks encourages the transformer-based models to work in tandem, with the first model identifying the relevant text segments and the second model accurately classifying the propaganda techniques employed within them. The best results obtained an F1 score of 62 and made use of a pre-trained RoBERTa model. This performance level suggests that accurately distinguishing between numerous, often subtly employed techniques remained a significant challenge for models of that generation, highlighting the potential for newer, larger language models and different methodologies like prompting to achieve further improvements.

In 2023, \cite{Mona} achieved state-of-the-art performance on the Qprop dataset. The authors employed a diverse range of models, including BERT, ELMo, LSA, FastText, and Word2Vec, obtaining an impressive F1 score of 97.76.

In \cite{Sprenkamp}, the authors used LLMs on the SemEval-2020 task 11 dataset. They combined multiple LLMs models such as: GPT-3.5 and GPT-4, and also a fine-tuned GPT-3.5 model, cross-joined with two prompting techniques: few-shot learning (base) and chain-of-thought learning (CoT). The obtained results were comparable to the ones obtained in the SemEval-2020 task 11, with an F1 score of 58.11 using GPT-4 model with few-shot-learning.

\subsection{Contribution}
We establish new state-of-the-art performance on the SemEval-2020 Task 11 Technique Classification subtask. Our fine-tuned DeBERTa V3 Large model achieves a macro F1-score of 63.18, surpassing the original SemEval winning score. Furthermore, our experiments with Causal LLMs show even higher performance, with Gemini Flash 2.5 reaching a macro F1-score of 63.62 using base prompting.

\section{Data}

\begin{figure}[t]
  \includegraphics[width=0.9\columnwidth]{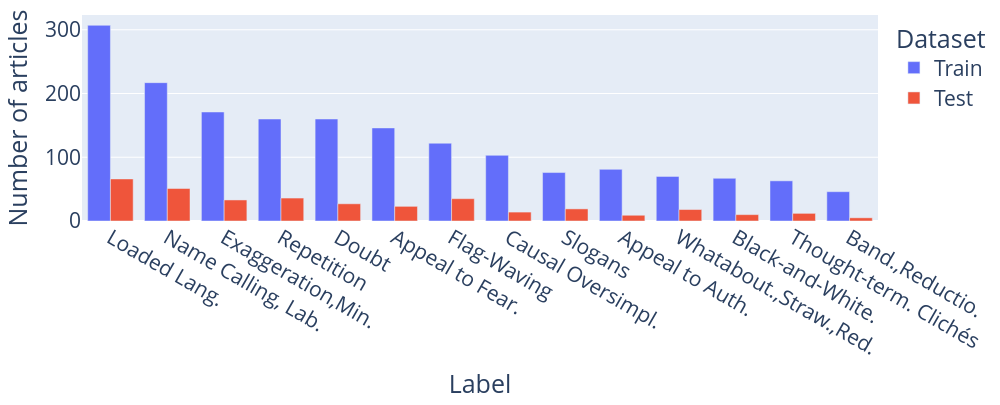}
  \caption{Labels distribution in the corpus, split by train and test. On the y-axis the number of articles and on the x-axis the labels. Loaded language is the most frequent technique. }
  \label{fig:data}
\end{figure}

We selected the SemEval-2020 Task 11 corpus \cite{da-san-martino-etal-2020-semeval} due to its comprehensiveness. This dataset offers a rich collection of examples, making it well-suited for training and evaluating our models. It was constructed using news articles sourced between mid-2017 and early 2019 from media outlets categorized by Media Bias/Fact Check. These articles underwent deduplication prior to annotation. The corpus comprises 536 articles. These articles are relatively lengthy, averaging 5,348 characters and 875 tokens per piece. Across these articles, there are 8,981 annotated propaganda instances. The quality of these annotations was assured by the collaborative efforts of six professional annotators, trained for this specific task. They employed a two-phase annotation protocol using a customized Anafora annotation platform \cite{chen-styler-2013-anafora}: initial independent annotation followed by consolidation with an expert reviewer. The reported inter-annotator agreement reached an average $\gamma$ score of $0.6$. The dataset is divided into standard training, development, and test splits, containing 371, 75, and 90 articles, respectively. The distribution of propaganda techniques is notably skewed, with \textit{Loaded Language} and \textit{Name Calling/Labeling} being the most frequent by a significant margin. Further details on the distribution of techniques and data splits can be seen in \autoref{fig:data}. The richness, annotation quality, and defined structure of the corpus make it an excellent resource for our investigation.

\section{Methodology}
The original SemEval task comprised two distinct but related subtasks: Span Identification (SI), which aimed to pinpoint the exact text segments (spans) within an article that contained propaganda, and Technique Classification (TC). The TC subtask, as originally conceived, involved taking these pre-identified spans and assigning one or more labels from a set of 14 predefined propaganda techniques (such as Loaded Language, Name Calling/Labeling, Appeal to Fear, etc.) to classify the specific method used within that particular span.

However, our research adapts this task. Instead of relying on pre-identified spans, we approach TC as an article-level, multi-label classification problem. We provide the entire text of each news article as input to our models and task them with identifying all propaganda techniques present anywhere within that article from the same list of 14 defined by the SemEval organizers. This approach tests the models' ability to discern the presence of various, potentially subtle, techniques embedded within the broader context of a full news piece. To achieve this, we employed and compared two primary categories of models: fine-tuned Masked Language Models (MLMs), specifically variants of RoBERTa and DeBERTa, and various prompted Causal Language Models (LLMs) from providers like Google, OpenAI, Anthropic, Mistral AI, and Meta.

\subsection{Masked Language Models}
In our first approach, we leveraged two state-of-the-art pre-trained language models, XLM-RoBERTa \cite{conneau2020unsupervised} and DeBERTa V3 Large \cite{he2021deberta}  and we fine-tuned them on the training dataset. On top of the pre-trained model, we put a fully connected layer composed of a linear layer, normalization, tanh activation function, and a 0.5 dropout layer. To optimize model performance, we adopted a two-phase fine-tuning strategy inspired by \cite{sun2020finetune}. In the first phase, we kept the original model's parameters frozen to preserve the pre-trained knowledge and solely trained the FC layer. This training was conducted over 5 epochs to allow the classifier to adapt to the domain-specific data without altering the core representations learned during pre-training. In the second phase, we unfroze the pre-trained model's final layer and jointly fine-tuned both the classification head and the last transformer layer of the base model. To mitigate the risk of overfitting, we applied a reduced learning rate ($2e-4$) during this phase and trained for one additional epoch.

\subsection{Causal Models}

For the Causal Model approach, we leveraged the generative capabilities of several state-of-the-art large language models (LLMs) accessed via APIs. The models evaluated include:
\begin{itemize}
\item Google: Gemini 1.0 Pro, Gemini 1.5 Pro, Gemini Flash 2.0, Gemini Flash 2.5
\item OpenAI: GPT-3.5-Turbo, GPT-4, GPT-4-Turbo, GPT-4o, GPT-4o-mini, a fine-tuned GPT-3.5-Turbo variant (details in \autoref{table:model_performance_base})
\item Anthropic: Claude 3 Haiku, Claude 3 Sonnet, Claude 3 Opus
\item Mistral AI: Mistral-Small (mistral-s), Mistral-Medium (mistral-m), Mistral-Large (mistral-l)
\item Meta: Llama 3
\end{itemize}

We interacted with these models using two distinct prompting strategies to perform the article-level multi-label classification task:

\begin{enumerate}
    \item \textbf{Base prompting}. We ask the conversational agent to return a list of identified propaganda detection techniques with the following prompt: 
    
    "For the given article state which of the 14 propaganda techniques are present. If no propaganda technique was identified, return 'no propaganda detected'. You MUST output a list of the propaganda techniques with each technique in a new line, e.g.: {techniques}. This is the article: <article>". 
    
    \item \textbf{Chain-of-thought prompting}. We ask conversational agent to return a list of identified propaganda detection techniques along with the explanation (reasoning) of choosing that technique with the following prompt: 
    
    "For the given article please state which of the 14 propaganda techniques are present and give an explanation to why the technique is present in the article. If no propaganda technique was identified return 'no propaganda detected'. You MUST output a list of the propaganda techniques with each technique in a new line, e.g.: Loaded Language - Your explanation why this technique is present in the article. This is the article: <article>".

\end{enumerate}

\section{Results}
Our investigation establishes new benchmarks for the SemEval-2020 Task 11 Technique Classification subtask. We obtained new SOTA results for both Masked Models and Causal Models. With our model based on XLM-RoBERTa we obtained an F1 score of 62.33, while the model based on DeBERTa V3 Large obtained a slightly better score of 63.18.

The best Causal Model (Gemini Flash 2.5) also beat the previous SOTA (63.62 vs 58). Using base prompting (as seen in \autoref{table:model_performance_base}) the models from Google performed best. Gemini Flash 2.5 has the highest recall (78) but a relatively lower precision (54), while Gemini Flash 2 has a more balanced precision and recall with a higher precision (61) than Gemini Flash 2.5.

Comparing prompting strategies for Causal LLMs yielded interesting insights. While Chain-of-Thought (CoT) prompting, designed to elicit reasoning, occasionally boosted recall significantly (as seen with GPT-4.1 and GPT-4o in \autoref{table:model_performance_chain}), it did not consistently lead to the highest F1 scores across all models. In fact, the top-performing model, Gemini Flash 2.5, achieved its best result using the simpler base prompt (\autoref{table:model_performance_base}). This indicates a trade-off: CoT might encourage models to identify more potential instances (high recall), but sometimes at the cost of accuracy (lower precision), potentially due to overthinking or generating less direct classifications. Furthermore, the performance tables reveal distinct behavioral profiles; for example, Gemini Flash 2.5 (base) excelled in recall (77.93) but had moderate precision (53.75), whereas our fine-tuned GPT-3.5 variant achieved high precision (71.68) but much lower recall (45.25).

Using chain-of-thought prompting (\autoref{table:model_performance_chain}), the top model is also Gemini Flash 2.5. Our results demonstrate that newer language models generally outperform older ones and that proprietary models outperform open source ones (LLaMA 3). This suggests that models are getting better at propaganda detection. Additionally, fine-tuning these models on specific datasets further enhances their performance compared to their base versions. Our fine-tuned version of gpt 3.5-turbo was better than the base one.

\begin{table}
    \caption{Model performance with Base Prompting (gpt-3.5-ft is a gpt-3.5-turbo fine-tuned model). The fine-tuned model was trained using 7,178,874 tokens over 3 epochs with a batch size of 2, resulting in 2246 steps, an LR multiplier of 2 and a seed of 477055734 (id: ft:gpt-3.5-turbo-0125:personal::9K750VBH).}
    \label{table:model_performance_base}
    \begin{tabular}{lccc}
    \hline
        \textbf{Model} & \textbf{F1-Score} & \textbf{Precision} & \textbf{Recall} \\
    \hline
        Gemini Flash 2.5 & 63.62 & 53.75 & 77.93 \\
        Gemini Flash 2.0 & 63.10 & 60.51 & 65.92 \\
        GPT-4.1 & 60.96 & 53.60 & 70.67 \\
        GPT-4o & 59.44 & 49.91 & 73.46 \\
        GPT-4o-mini & 57.67 & 51.42 & 65.64 \\
        gpt-3.5-ft & 55.48 & 71.68 & 45.25 \\
        mistral-s & 52.27 & 44.06 & 64.25 \\
        claude-sonnet & 51.94 & 58.60 & 46.65 \\
        gpt-4-turbo & 51.64 & 52.77 & 50.56 \\
        claude-opus & 50.17 & 63.79 & 41.34 \\
        mistral-m & 46.75 & 48.36 & 45.25 \\
        Gemini Pro 1.5 & 46.28 & 44.88 & 47.77 \\
        claude-haiku & 42.75 & 47.14 & 39.11 \\
        gpt-3.5-turbo & 42.65 & 59.50 & 33.24 \\
        mistral-l & 40.35 & 42.68 & 38.27 \\
        Gemini Pro 1.0 & 37.15 & 45.71 & 31.28 \\
        llama3 & 27.38 & 42.86 & 20.11 \\
        \hline
    \end{tabular}
\end{table}

\begin{table}
    \caption{Model performance with Chain-of-thought prompting. Best performing model is Gemini Flash 2.5, but the model with the highest recall was GPT-4o.}
    \label{table:model_performance_chain}
    \begin{tabular}{lccc}
    \hline
        \textbf{Model} & \textbf{F1-Score} & \textbf{Precision} & \textbf{Recall} \\
    \hline
        Gemini Flash 2.5 & 61.55 & 50.63 & 78.49 \\
        Gemini Flash 2.0 & 59.77 & 55.05 & 65.36 \\
        GPT-4o-mini & 58.67 & 51.82 & 67.60 \\
        GPT-4o & 56.58 & 43.64 & 80.45 \\
        gpt-4-turbo & 55.31 & 52.10 & 58.94 \\
        GPT-4.1 & 52.88 & 36.68 & 94.69 \\
        claude-opus & 50.39 & 56.40 & 45.53 \\
        mistral-s & 50.23 & 37.31 & 76.82 \\
        claude-sonnet & 49.13 & 51.21 & 47.21 \\
        Gemini Pro 1.5 & 46.39 & 50.33 & 43.02 \\
        mistral-m & 45.06 & 42.64 & 47.77 \\
        mistral-l & 40.46 & 41.59 & 39.39 \\
        claude-haiku & 36.86 & 43.23 & 32.12 \\
        Gemini Pro 1.0 & 35.53 & 40.65 & 31.56 \\
        gpt-3.5-turbo & 31.93 & 46.52 & 24.30 \\
        llama3 & 25.61 & 39.31 & 18.99 \\
    \hline
    \end{tabular}
\end{table}

\subsection{Error Analysis and Discussion}

We can see in \autoref{table:model_performance_by_label} that there's a significant difference in performance across different models and techniques. Some models excel at certain techniques while struggling with others. Certain propaganda techniques appear to be more challenging to detect than others. Most model fare well in the loaded language and name calling techniques. Models struggle in general with cliches, bandwagon and black and white fallacy. We observe substantial performance disparities among models for the same propaganda technique. For instance, Claude-Sonnet achieves a notably higher accuracy of 0.89 for "loaded language" compared to GPT-3.5's 0.49. This suggests that an ensemble approach, combining multiple models, could yield improved results due to their complementary strengths.

\begin{table*}[htbp]
\centering
\begin{tabular}{l|c|c|c|c|c|c|c|c|c|c|c}
\hline
 \thead{Techniques} & \thead{Claude\\ sonnet} & \thead{OpenAI\\ gpt-3.5} & \thead{OpenAI\\ GPT-4o} & \thead{OpenAI\\ gpt-4} & \thead{Meta\\ llama3} & \thead{Mistral\\ large} & \thead{Mistral\\ medium} & \thead{Gemini\\ 1.0-pro} & \thead{Gemini \\1.5-pro} & \thead{Gemini \\Flash 2.0} & \thead{Gemini \\Flash 2.5}  \\ \hline
 appeal\_to\_authority  & 0.38 & 0.00 & 0.37 & 0.00 & 0.14 & 0.36 & 0.21 & 0.10 & 0.43 & 0.29 & 0.33 \\
 appeal\_to\_fear   & 0.53 & 0.35 & 0.48 & 0.49 & 0.07 & 0.45 & 0.29  & 0.51 & 0.62 & 0.63 & 0.64 \\
 bandwagon  & 0.12 & 0.00 & 0.16 & 0.21 & 0.10 & 0.00 & 0.00  & 0.09 & 0.14 & 0.18 & 0.18 \\
 black\_and\_white  & 0.13 & 0.00 & 0.35 & 0.35 & 0.00 & 0.30 & 0.31 & 0.22 & 0.26 & 0.47 & 0.29 \\
 oversimplification & 0.29 & 0.00 & 0.31 & 0.41 & 0.00 & 0.32 & 0.37 & 0.27 & 0.36 & 0.41 & 0.46 \\
 doubt  & 0.35 & 0.31 & 0.30 & 0.54 & 0.00 & 0.48 & 0.59  & 0.38 & 0.51 & 0.52 & 0.56 \\
 exaggeration  & 0.34 & 0.11 & 0.68 & 0.62 & 0.11 & 0.52 & 0.39  & 0.05 & 0.06 & 0.61 & 0.66 \\
 flag\_waving  & 0.05 & 0.33 & 0.59 & 0.56 & 0.00 & 0.05 & 0.44 & 0.20 & 0.25 & 0.39 & 0.69 \\
 loaded\_language & 0.89 & 0.49 & 0.95 & 0.95 & 0.84 & 0.71 & 0.94  & 0.53 & 0.88 & 0.95 & 0.94 \\
 name\_calling & 0.70 & 0.61 & 0.82 & 0.58 & 0.07 & 0.49 & 0.04  & 0.64 & 0.57 & 0.77 & 0.82 \\
 repetition & 0.32 & 0.15 & 0.66 & 0.46 & 0.12 & 0.19 & 0.60  & 0.37 & 0.24 & 0.64 & 0.63 \\
 slogans & 0.41 & 0.00 & 0.51 & 0.00 & 0.08 & 0.14 & 0.33  & 0.13 & 0.29 & 0.24 & 0.44 \\
 cliches  & 0.15 & 0.00 & 0.30 & 0.29 & 0.00 & 0.00 & 0.13  & 0.32 & 0.13 & 0.00 & 0.24 \\
 whataboutism & 0.41 & 0.38 & 0.42 & 0.39 & 0.09 & 0.41 & 0.32 & 0.24 & 0.21 & 0.33 & 0.43 \\
\hline
\end{tabular}
\caption{Comparative analysis of causal language models' performance (F1 macro scores) in detecting specific propaganda techniques.}
\label{table:model_performance_by_label}
\end{table*}

\begin{figure}[t]
  \includegraphics[width=1\columnwidth]{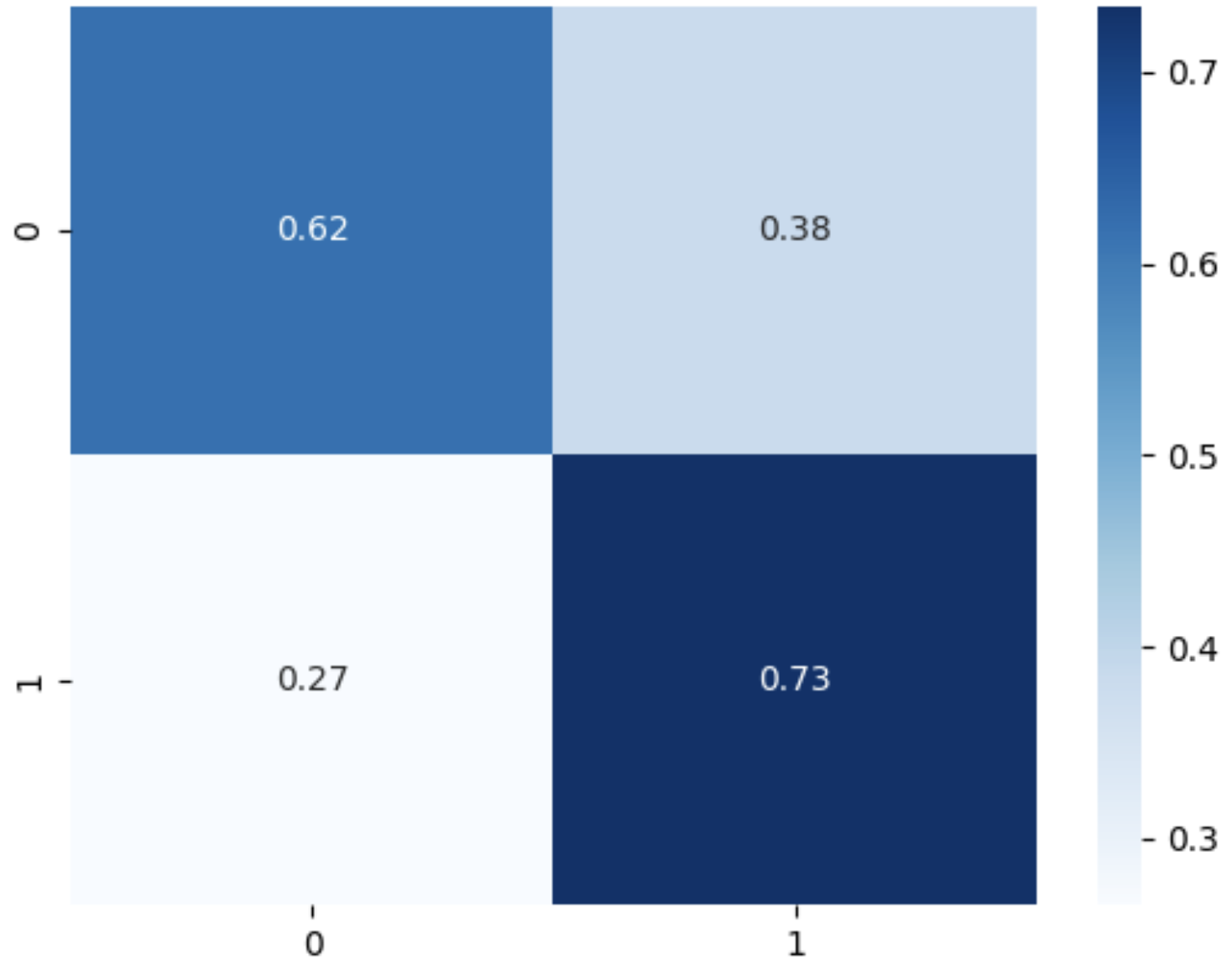}
  \caption{This confusion matrix visualizes the performance for the model GPT-4o, using base prompting. (class 0 is the existence of propaganda technique and class 1 is the absence of propaganda technique) }
  \label{fig:cf_base}
\end{figure}

If we average the F1 score of each technique we obtain an overall F1 score of 65.61. Based on the confusion matrix of results with base prompting in \autoref{fig:cf_base}, the model performs better on class 1 (with a 73\% true positive rate) than on class 0 (with a 62\% true negative rate). False positives (38\%) are a more common error than false negatives (27\%), indicating that the model tends to over-predict class 1 slightly more often than class 0.

\begin{figure}[t]
  \includegraphics[width=1\columnwidth]{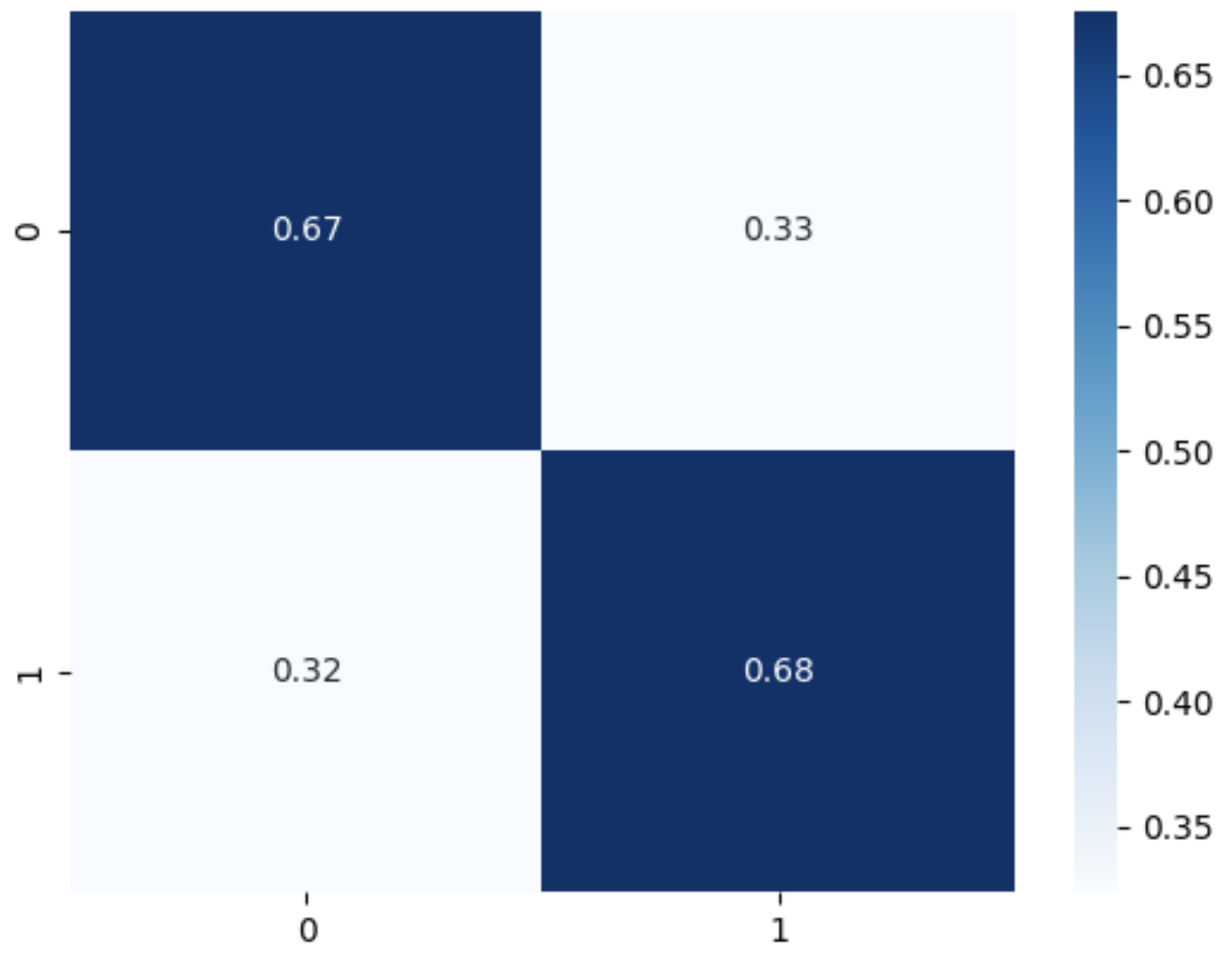}
  \caption{This confusion matrix visualizes the performance for the model GPT-4o-mini, using chain-of-thought prompting.}
  \label{fig:cf_chain}
\end{figure}

Looking at the same confusion matrix for chain-of-thought prompting (\autoref{fig:cf_chain}), we have an average F1 score of 67.34, reflecting a slight improvement over the base prompting. The model performs similarly on both classes, with slightly better results for class 0 (67\% true negative) than class 1 (68\% true positive).

\subsection{Conclusions and Future Work}

In this paper, we explored advancements in propaganda detection through a comparative analysis of masked and causal language models, leveraging base and chain-of-thought prompting strategies. Our results demonstrate that newer language models, particularly causal models like GPT-4o and Gemini Flash 2.0 and 2.5, outperform previous state-of-the-art approaches in identifying propaganda techniques. Our findings suggest that fine-tuning models on specific datasets and employing diverse prompting techniques, such as chain-of-thought prompting, further improve detection accuracy. Also, when using Masked Models, fine-tuning them in 2 steps and freezing the weights improve the model's performance.

For future work, we recommend exploring the use of ensemble models. As highlighted in the error analysis, different models exhibit strengths in detecting distinct propaganda techniques. Future work could focus on developing ensemble methods that combine predictions from multiple top-performing models (both masked and causal) to leverage their complementary capabilities and potentially achieve more robust and accurate overall detection across all techniques. We would also like to investigate hybrid models that combine the strengths of fine-tuned Masked Language Models (like DeBERTa V3 for its strong baseline performance) with the contextual understanding and reasoning abilities of large Causal Models (like Gemini or GPT-4) perhaps using the MLM output to guide or constrain the LLM's analysis.

This study focused solely on Technique Classification (TC). Future research could revisit the full SemEval task by integrating Span Identification (SI) with TC using modern LLMs. This could involve multi-task learning or pipeline approaches where LLMs first identify potential spans and then classify the techniques within those specific spans, potentially improving precision. Also, given that propaganda dissemination is a global phenomenon transcending language barriers, future work should extend this research beyond English to evaluate model performance in other languages and develop robust multilingual detection systems.

\subsection{Limitations}
Evaluating Large Language Models faces the challenge of potential data leakage. Because these models are pre-trained on vast datasets, scraped from the internet, there's a risk they were exposed to the test data—in this instance, the SemEval task data—during training, which could artificially inflate performance and obscure true generalization ability.

\subsection{Ethical considerations}
There are no ethical issues that could result from the publication of our work.

\begin{acks}

This research is partially supported by the project “Romanian Hub for Artificial Intelligence - HRIA”, Smart Growth, Digitization and Financial Instruments Program, 2021-2027, MySMIS no. 334906.
\end{acks}


\bibliographystyle{ACM-Reference-Format}
\bibliography{sample-base}

\end{document}